\documentclass[letterpaper]{article} % DO NOT CHANGE THIS
\usepackage[preprint]{aaai2027}  % DO NOT CHANGE THIS
\usepackage[hyphens]{url}  % DO NOT CHANGE THIS
\usepackage{graphicx} % DO NOT CHANGE THIS
\usepackage{natbib}  % DO NOT CHANGE THIS AND DO NOT ADD ANY OPTIONS TO IT
\usepackage{caption} % DO NOT CHANGE THIS AND DO NOT ADD ANY OPTIONS TO IT
\usepackage{algorithm}
\usepackage{algorithmic}

\usepackage{newfloat}
\usepackage{listings}
\usepackage{tabularray}
\usepackage{multirow}
\DeclareCaptionStyle{ruled}{labelfont=normalfont,labelsep=colon,strut=off} % DO NOT CHANGE THIS
\floatstyle{ruled}
\newfloat{listing}{tb}{lst}{}
\floatname{listing}{Listing}

\usepackage{booktabs}

\title{Source-Face Authenticity Detection for 3D Gaussian Heads Reconstructed from a Single Portrait: A Benchmark and Dedicated Detector}
\author{
    Yujie Gao\textsuperscript{\rm 1}, Zijian Yu\textsuperscript{\rm 2}, Yan Hong\textsuperscript{\rm 2}, Jun Lan\textsuperscript{\rm 2},
    Jianfu Zhang\textsuperscript{\rm 1}\corresponding
}
\affiliations{
    \textsuperscript{\rm 1}Shanghai Jiaotong University \textsuperscript{\rm 2}Ant Group\\

    \textsuperscript{\texttt{GYJgyj123@sjtu.edu.cn; thss15\_yuzj@163.com; yanhong.sjtu@gmail.com; lanjun\_yelan@163.com; c.sis@sjtu.edu.cn}}
    
}

\begin{document}

\maketitle

\begin{abstract}
Recent advances in single-image 3D Gaussian head reconstruction have enabled highly realistic and freely renderable digital heads from a single portrait. However, reconstruction and rendering can weaken the forgery traces in the source portrait, making the resulting 3D face difficult to classify whether its underlying face is real or fake, and thereby posing risks to identity authentication and face privacy.
To study this problem, we introduce the first large-scale benchmark for this task by collecting real portraits and fake portraits from multiple sources and evaluate representative existing detectors on this benchmark, revealing their lack of explicit mechanisms for retaining fine-grained information and maintaining feature consistency across rendered views. To directly address these two limitations, we propose a detector trained with a two-stage strategy. In Stage I, masked autoencoding encourages the visual backbone to retain the fine-grained appearance information required for local reconstruction, while multi-view contrastive learning enforces feature consistency across rendered views of the same head. Since CLS tokens at different depths exhibit complementary spatial attention patterns, Stage II freezes the adapted backbone and concatenates low-, middle-, and high-level CLS tokens for classification. Experiments show that our method achieves the highest accuracy and ranks first across all reported metrics among the evaluated detectors.

\end{abstract}


\section{Introduction}
Recent advances in single-image 3D Gaussian head generation have made it increasingly practical to recover photorealistic and renderable 3D heads from only one portrait, enabling applications in digital humans, telepresence, content creation, and AR/VR. Yet a reconstructed 3D head is also an identity-bearing asset: when generated from a forged or manipulated portrait, a 3D Gaussian head may provide a persistent, freely renderable identity asset that can facilitate impersonation or attacks on identity-authentication systems. This makes authenticity detection essential for protecting both identity authentication and face privacy in 3D face applications. However, existing face forgery and fake image detectors mainly target 2D images or videos, and no dedicated method has yet been developed to determine, from a single rendered view of a 3D Gaussian head, whether its underlying source face is real or fake.

\begin{figure}
    \centering
    \includegraphics[width=1\linewidth]{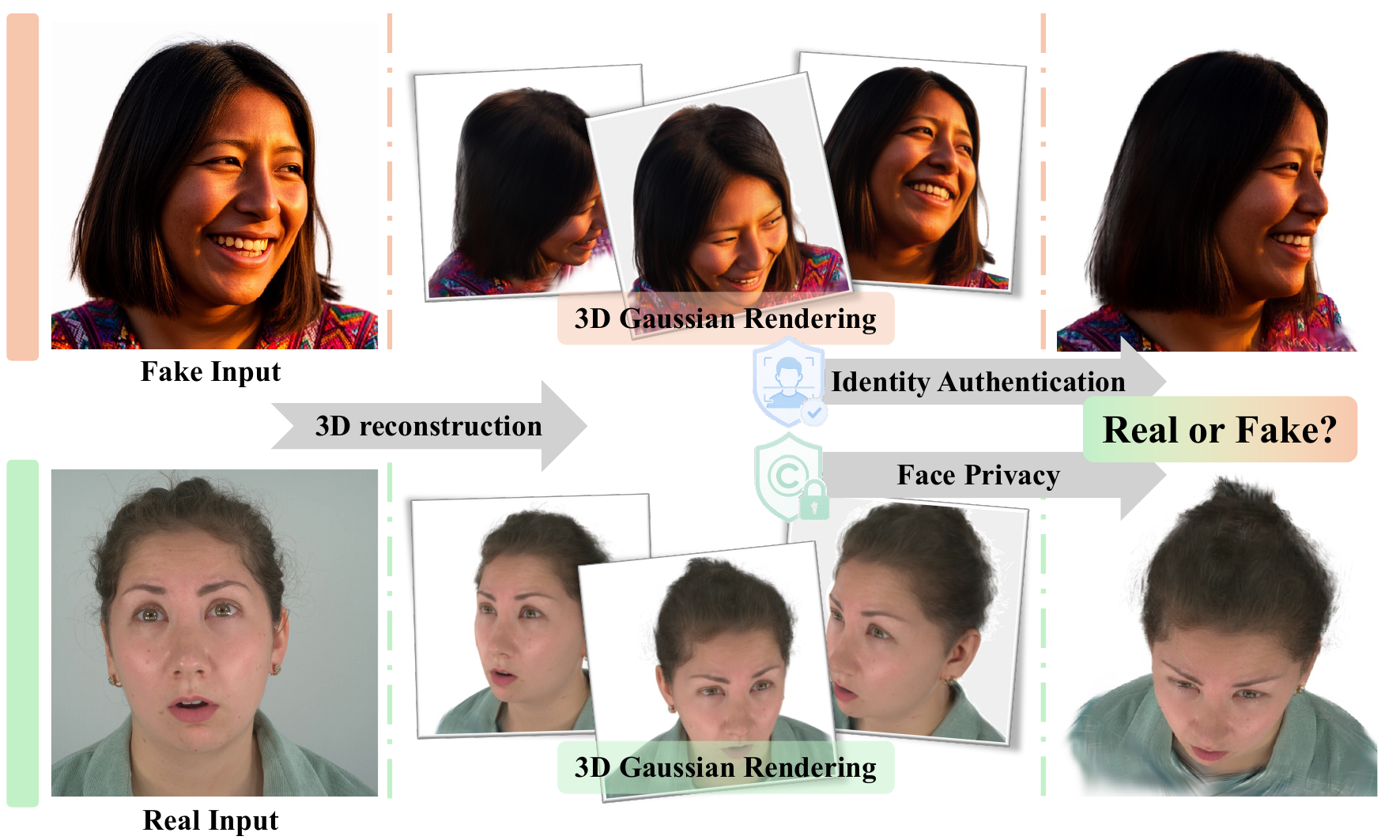}
    \caption{\textbf{Task definition.} Given a 3D Gaussian head reconstructed from a single portrait image, our task is to determine whether its underlying face is real or fake. At inference time, the authenticity decision is made from a single rendered view of the reconstructed head.}
    \label{fig:task-definition}
\end{figure}

The evolution of single-image 3D head generation reflects a steady move toward more expressive and realistic representations. Early methods typically relied on fitting parametric face models, such as 3D morphable models~\cite{3DMM} and FLAME~\cite{FLAME}, to estimate facial shape, expression, and pose from an input image. Later neural approaches introduced more flexible implicit 3D representations, including tri-plane features~\cite{panohead,rodinhd,spherehead} and NeRF-based radiance fields~\cite{nerface}. More recently, 3D Gaussian Splatting~\cite{3dgs} has become a promising representation for head reconstruction. By optimizing dense Gaussians from a single portrait, recent 3D Gaussian head methods~\cite{cap4d,facelift,lam,any3davatar} can generate increasingly photorealistic heads that faithfully preserve the identity, geometry, expression, and texture cues of the input image across novel views, which also poses new challenges for identity authentication and face privacy protection.

\begin{figure}
    \centering
    \includegraphics[width=1\linewidth]{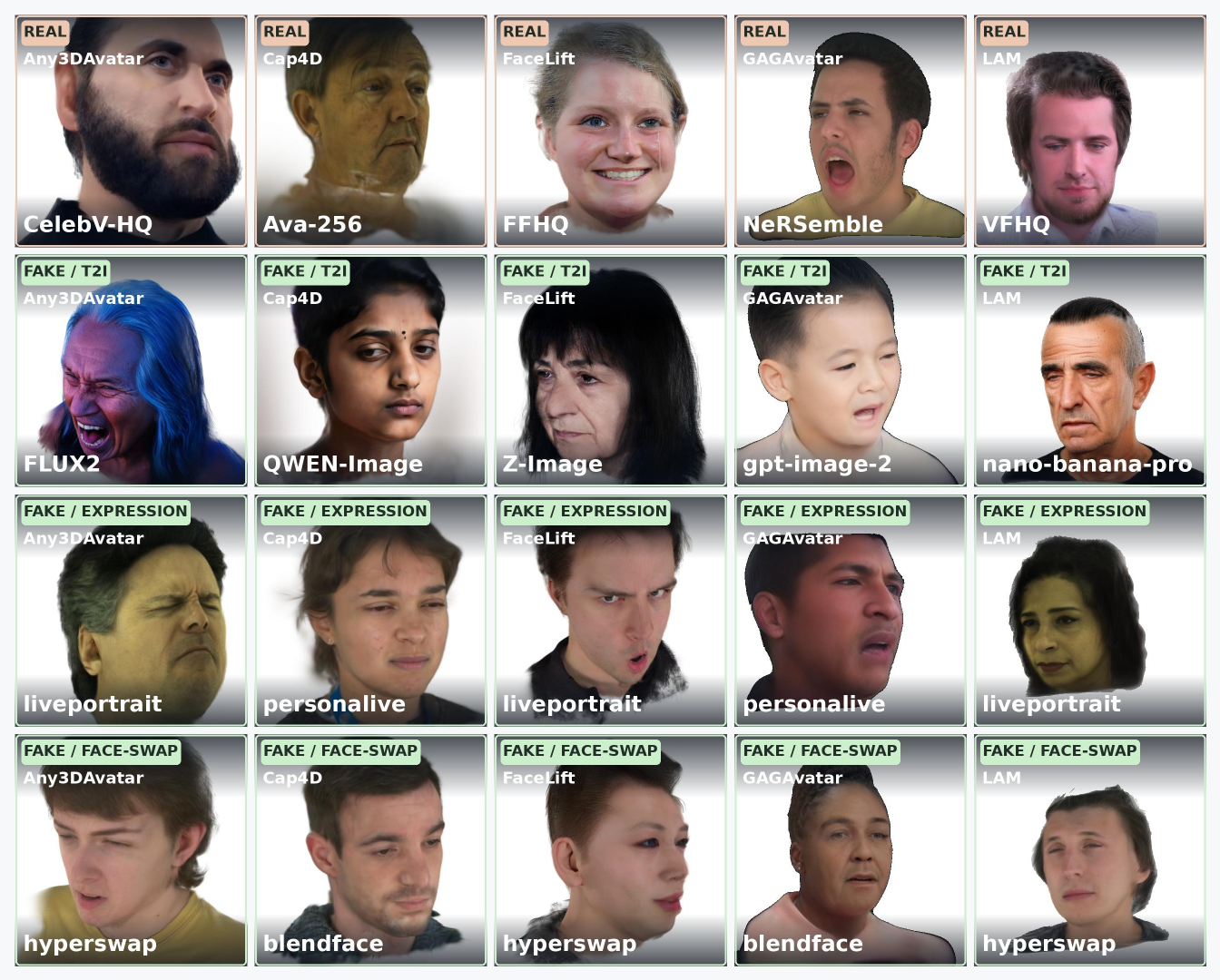}
    \caption{Overview of our 3D Gaussian head dataset.}
    \label{fig:dataset-overview}
\end{figure}

In parallel, 2D visual authenticity detection has been extensively studied for both face forgery and general fake-image detection~\cite{iid,faceforensics++,ForensicsAdapter,npr,pgc,effort,freqnet}. Existing methods primarily distinguish real and fake content using manipulation artifacts, texture inconsistencies, generator fingerprints, or statistical cues in 2D images and videos. However, directly applying these 2D detectors to authenticity detection for 3D Gaussian heads reconstructed from a single image presents substantial challenges: (1) 3D Gaussian reconstruction and re-rendering may compress or alter the original 2D forgery cues, leaving only subtle, localized evidence. Existing detectors lack explicit mechanisms to retain such fine-grained information, making these cues difficult to capture. (2) Changes in rendering viewpoint alter the visibility, scale, and projection of these cues, yielding different local evidence patterns across views of the same 3D Gaussian head. As a result, detectors trained only with standard 2D supervision may overfit to view-specific rendering shortcuts rather than learning view-robust authenticity cues. These challenges motivate a detector that preserves fine-grained appearance information in the 3DGS rendering domain and learns consistent authenticity representations across rendered views, even when only a single view is used at inference time.

To systematically study this problem, we make three main contributions: (1) We introduce, to the best of our knowledge, \textbf{the first large-scale dataset for real/fake 3D Gaussian head detection}, where 2D forged faces from 9 methods across text-to-image generation, face swapping, and expression-driven synthesis, together with genuine portraits from 5 real-face datasets, are reconstructed by 5 single-image 3D head methods and rendered into approximately 360K multi-view 3DGS head samples; (2) \textbf{We establish a benchmark on this dataset} and evaluate representative baselines from both face forgery detection and general fake image detection, revealing their limitations in retaining fine-grained authenticity cues after Gaussian reconstruction and maintaining feature consistency across rendered views; (3) We show that reliable 3DGS head detection benefits from preserving fine-grained appearance information and enforcing cross-view feature consistency, while aggregating regional cues across network depths further improves classification. Guided by these findings, \textbf{we propose a two-stage detector}: Stage I combines masked autoencoding~\cite{mae} for detail-preserving representation learning with multi-view contrastive learning for cross-view consistency, while Stage II freezes the adapted backbone and fuses low-, middle-, and high-level CLS tokens for single-view classification. 

In summary, our contributions are:
\begin{itemize}
    \item We introduce the first large-scale dataset for real/fake 3D Gaussian head detection, comprising approximately 360K multi-view renderings.
    \item We establish a systematic benchmark with balanced, identity-disjoint splits and evaluate representative recent 2D authenticity detectors in the 3DGS head domain under a unified protocol.
    \item We propose a two-stage detector that combines masked autoencoding and multi-view contrastive learning for representation learning, followed by multi-level CLS token fusion for classification.
\end{itemize}

% \begin{figure}
%     \centering
%     \includegraphics[width=1\linewidth]{Images/source_by_reconstruction_heatmap.pdf}
%     \caption{}
%     \label{fig:distrubution}
% \end{figure}

\section{Related Works}

\noindent\textbf{3D Gaussian Head Reconstruction from Single Image: }
3D Gaussian head reconstruction from a single portrait image has recently become an active research direction. Compared with earlier mesh- or implicit-field-based representations, 3DGS provides an explicit and efficiently renderable representation that can faithfully reproduce the input view while plausibly completing invisible regions of the head. Existing methods can be roughly grouped into several lines. Some approaches adopt a two-stage pipeline, first generating multi-view observations from the input portrait and then reconstructing a Gaussian point cloud from these synthesized views, as exemplified by FaceLift~\cite{facelift}. Another line incorporates parametric head priors, especially FLAME, into Gaussian representations to improve geometric stability, animation controllability, and reconstruction quality, including LAM~\cite{lam}, CAP4D~\cite{cap4d}, and GAGAvatar~\cite{gagavatar}. More recent efforts further pursue one-step and efficient generation, aiming to reduce reconstruction latency while preserving high-quality full-head appearance, such as FastAvatar~\cite{fastavatar} and Any3DAvatar~\cite{any3davatar}. These advances make single-image 3DGS head reconstruction increasingly practical, but they also raise the need to verify whether the reconstructed head originates from a real or fake face. Recent work such as Fake3DGS~\cite{fake3dgs, gs-checker} has explored detecting manipulations of 3DGS representations, but it focuses on tampering with the Gaussian point cloud rather than the authenticity of the underlying source face and is therefore not directly applicable to our task.

\begin{figure*}[t]
    \centering
    \includegraphics[width=1\linewidth]{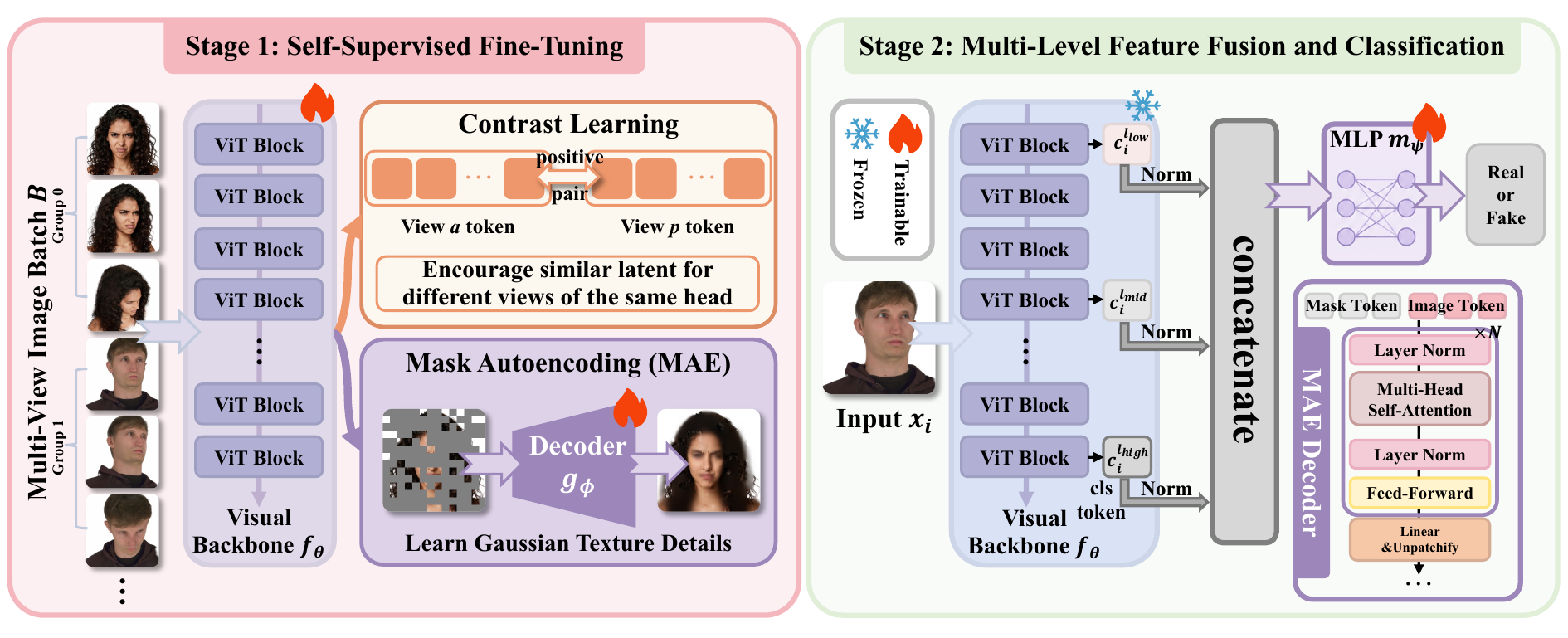}
    \caption{\textbf{Overview of our two-stage training framework.} In Stage 1, we train the visual backbone with two self-supervised objectives: masked autoencoding models local texture details and subtle Gaussian artifacts, while multi-view contrastive learning suppresses view-specific shortcuts. In Stage 2, low-, middle-, and high-level CLS tokens are normalized and concatenated, then fed into an MLP classification head for real/fake prediction.}
    \label{fig:pipeline}
\end{figure*}
\noindent\textbf{AI-Generated Face Detection: }
Contemporary face forgeries mainly arise from three categories: text-to-image face generation~\cite{qwen-image,z-image,flux2,nanobanana,gptimage2}, expression-driven face reenactment~\cite{liveportrait,personalive,xnemo}, and face swapping~\cite{blendface,ghost,hififace,simswap}. Accordingly, existing detection methods have developed along two major directions. The first is tailored specifically to facial content and exploits face-specific forensic evidence, such as identity inconsistency, blending boundaries, and localized manipulation artifacts, to distinguish forged faces from genuine ones~\cite{iid,faceforensics++,ForensicsAdapter,tall,rfm,Multi-attention, SIA, LSDA,DSI}. The second treats forged faces as a special case of general AI-generated imagery and develops detectors over broader image domains~\cite{adie,effort,pgc,freqnet,dda,drct}. By learning from diverse semantic content and generation pipelines, these general-purpose methods can capture a wider spectrum of forgery traces and often exhibit stronger cross-generator generalization. Nevertheless, both directions are designed for native 2D images, leaving the authenticity detection of faces reconstructed and rendered as 3D Gaussian heads largely unexplored.

\section{Methodology}

\subsection{Dataset Construction}
Recent single-image 3D Gaussian head reconstruction methods~\cite{cap4d,facelift,lam,gagavatar,any3davatar,fastavatar,arc2avatar} can generate high-fidelity 3D heads that nearly replicate the input portrait, particularly when rendered from viewpoints close to the input view. Consequently, forged or manipulated portraits can become persistent, freely renderable 3D identity assets, threatening identity authentication and intellectual property protection. To address this risk, we construct, to the best of our knowledge, the first large-scale dataset for real/fake 3D Gaussian head detection.

\noindent\textbf{Dataset Details: }
Our dataset contains 361,469 rendered images from 16,372 unique identities, comprising 170,060 real and 191,409 fake samples. Fig.~\ref{fig:dataset-overview} presents representative examples from the resulting dataset. We collect real portraits from 5 real-face datasets: CelebV-HQ~\cite{CelebV-HQ}, Ava-256~\cite{ava-256}, FFHQ~\cite{FFHQ}, NeRSemble~\cite{nersemble}, and VFHQ~\cite{VFHQ}. We employ 9 2D face forgery methods to construct the fake portraits. Both real and fake portraits are then reconstructed using 5 single-image 3D Gaussian head methods. Each reconstructed head is rendered at a resolution of $512\times512$ from 5--10 randomly sampled viewpoints, with yaw and pitch ranging from $-45^\circ$ to $45^\circ$ and the camera radius sampled from $0.8$ to $1.2$ relative to a normalized standard head scale.

\noindent\textbf{Fake Head Generators: }
To construct fake 3D heads, we first generate forged 2D portraits spanning three categories: text-to-image generation, face swapping, and expression-driven synthesis. For each category, we select a combination of recent and well-established forgery methods to cover both contemporary generation capabilities and representative classic pipelines. For text-to-image generation, we define 10 facial and appearance attributes, including gender, hairstyle, expression, and ethnicity, and prepare 50 candidate values for each attribute. We randomly sample one value from each attribute and combine the selected values into a prompt, which is then fed into Z-Image~\cite{z-image}, Qwen-Image~\cite{qwen-image}, GPT-Image-2~\cite{gptimage2}, Nano Banana Pro~\cite{nanobanana}, or FLUX.2~\cite{flux2}. For face swapping and expression-driven synthesis, we randomly sample two portraits from the same real face datasets used for the genuine subset. The two portraits serve as the source and target for face swapping with HyperSwapper~\cite{hyperswap} or BlendFace~\cite{blendface}, and as the source and driving reference for expression transfer with LivePortrait~\cite{liveportrait} or PersonaLive~\cite{personalive}. Drawing these inputs from the real data sources helps reduce identity-distribution bias between real and fake samples.
Each resulting forged portrait is subsequently reconstructed using one of five single-image 3D Gaussian head methods. We include both optimization-based and feed-forward pipelines: CAP4D~\cite{cap4d}, FaceLift~\cite{facelift}, LAM~\cite{lam}, GAGAvatar~\cite{gagavatar}, and Any3DAvatar~\cite{any3davatar}. The reconstructed heads are finally rendered from multiple randomly sampled viewpoints using the protocol described above to produce the fake 3D head samples.

\subsection{Two Stage Training}
Existing 2D detectors face two challenges. First, 3D Gaussian reconstruction and re-rendering weaken forgery cues, leaving only subtle local evidence. Second, viewpoint changes cause this evidence to vary across views of the same head. As shown in Fig.~\ref{fig:pipeline}, we address these challenges with a two-stage framework that first learns representations that preserve fine-grained appearance details and remain consistent across views, and then freezes the visual backbone and fuses low-, middle-, and high-level CLS tokens for authenticity classification. 

\noindent\textbf{Self-Supervised Training:}
Given an image batch $\mathcal{B}$, Each batch $\mathcal{B}=\{\mathcal{G}_g\}_{g=1}^{G}$ contains $G$ identity groups, where $\mathcal{G}_g=\{\mathbf{x}_{g,k}\}_{k=1}^{K}$ consists of $K$ views rendered from the same 3D head, while different groups correspond to different identities. The total batch size is $B=GK$. For the MAE branch, we flatten the grouped batch as $\{\mathbf{x}_i\}_{i=1}^{B}$ and divide each image into $P$ non-overlapping patches $\{\mathbf{p}_{i,j}\}_{j=1}^{P}$.

\textit{Masked Autoencoding.} To adapt the backbone to the 3DGS rendering domain and encourage it to retain fine-grained local appearance information, we randomly mask a ratio $\rho$ of the patches in the MAE branch, forming a masked index set $\mathcal{M}_i$ and a visible index set $\mathcal{V}_i$, where $|\mathcal{M}_i|=\rho P$ and $\mathcal{V}_i=\{1,\ldots,P\}\setminus\mathcal{M}_i$. Only the visible patches are embedded and passed through a ViT-style visual backbone $f_{\theta}$ to produce image tokens
\begin{equation}
\mathbf{Z}_i=f_{\theta}\left(\{\mathbf{p}_{i,j}\}_{j\in\mathcal{V}_i}\right).
\end{equation}
We restore the image tokens to their original patch positions, insert a learned mask token at each masked position, and add positional embeddings. The complete token sequence is then processed by an MAE decoder $g_{\phi}$ composed of $N$ ViT blocks. A linear layer maps each decoder output token to the flattened RGB pixels of its corresponding patch, yielding the prediction $\hat{\mathbf{p}}_{i,j}$. We reconstruct only the unseen patches using
\begin{equation}
\mathcal{L}_{\mathrm{MAE}}=\frac{1}{B\rho P}\sum_{i=1}^{B}\sum_{j\in\mathcal{M}_i}\left\|\hat{\mathbf{p}}_{i,j}-\mathbf{p}_{i,j}\right\|_2^2.
\label{eq:mae-loss}
\end{equation}
The reconstruction objective does not directly supervise authenticity labels or artifact locations. Instead, inferring missing content from visible patches provides target-domain self-supervision and encourages the backbone to preserve structural and fine-grained appearance information. We hypothesize that this information-preserving representation provides a stronger basis for the downstream real/fake classifier, a hypothesis evaluated by the component ablation in Experiment Section.

\textit{Multi-View Contrastive Learning.} Unlike the MAE branch, the contrastive branch feeds the complete patch sequence of each view into the visual backbone without masking. We extract and $\ell_2$-normalize its CLS token as the latent representation $\mathbf{z}_{g,k}$. Let $\mathcal{I}=\{(g,k)\}$ denote all views in the batch. For an anchor $a=(g,k)$, its positive set $\mathcal{P}(a)=\{(g,q)\mid q\neq k\}$ contains the other views in the same group, while $\mathcal{A}(a)=\mathcal{I}\setminus\{a\}$ contains all non-anchor views. We first define the normalized similarity between anchor $a$ and candidate $p$ as
\begin{equation}
q(a,p)=\frac{\exp(\mathbf{z}_a^{\top}\mathbf{z}_p/\tau)}{\sum_{n\in\mathcal{A}(a)}\exp(\mathbf{z}_a^{\top}\mathbf{z}_n/\tau)},
\label{eq:contrastive-similarity}
\end{equation}
where $\tau$ is the temperature. The multi-positive contrastive loss is
\begin{equation}
\mathcal{L}_{\mathrm{con}}=-\frac{1}{B}\sum_{a\in\mathcal{I}}\frac{1}{|\mathcal{P}(a)|}\sum_{p\in\mathcal{P}(a)}\log q(a,p).
\label{eq:contrastive-loss}
\end{equation}
This objective pulls views of the same head together in the latent space and pushes views from different identity groups apart, promoting viewpoint-invariant authenticity features.

\begin{table*}
\centering

\small
\setlength{\tabcolsep}{1mm}
\begin{tabular}{c|cccc|ccccc|c} 
\hline
\multirow{2}{*}{\textbf{Method}} & \multicolumn{4}{c|}{\textbf{2D Split }}                                  & \multicolumn{5}{c|}{\textbf{3D Split }}                                                         & \multirow{2}{*}{\textbf{Total}}  \\ 
\cline{2-10}
                                 & \textbf{T2I}   & \textbf{ExpDriven} & \textbf{FaceSwap} & \textbf{Real}  & \textbf{Any3DAvatar} & \textbf{Cap4D} & \textbf{FaceLift} & \textbf{GAGAvatar} & \textbf{LAM}   &                                  \\ 
\hline
\textbf{IID}                     & 89.32          & 54.04              & 71.72             & 85.05          & 83.60                & 77.97          & 77.20             & 73.00              & 80.10          & 78.60                            \\
\textbf{NPR}                     & 88.72          & 70.04              & 76.80             & 84.84          & 86.60                & 81.67          & 82.07             & 73.57              & 84.50          & 81.68                            \\
\textbf{FreqNet}                 & 83.48          & 48.64              & 54.64             & 79.04          & 76.83                & 71.80          & 67.07             & 65.63              & 71.90          & 70.65                            \\
\textbf{ForAda}                  & 95.76          & 58.28              & 79.20             & 89.15          & 89.37                & 79.80          & 84.13             & 78.20              & 85.73          & 82.49                            \\
\textbf{AIDE}                    & 95.12          & 56.68              & 68.60             & 72.56          & 75.37                & 71.93          & 72.17             & 72.27              & 73.33          & 73.01                            \\
\textbf{Effort}                  & 96.20          & 63.40              & 89.44             & 87.32          & 89.20                & 81.60          & 84.67             & 76.93              & 82.67          & 85.17                            \\
\textbf{PGC}                     & 96.04          & 63.96              & 83.88             & 83.05          & 84.67                & 82.33          & 81.10             & 76.87              & 85.90          & 82.17                            \\ 
\hline
\textbf{Ours}                    & \textbf{98.28} & \textbf{76.24}     & \textbf{90.04}    & \textbf{90.84} & \textbf{94.97}       & \textbf{84.10} & \textbf{91.67}    & \textbf{84.77}     & \textbf{92.07} & \textbf{89.51}                   \\
\hline
\end{tabular}
\caption{\textbf{Accuracy comparison across 2D forgery categories and 3D reconstruction methods.}}
\label{tab:split-results}
\end{table*}
\begin{table}
\centering

\small
\setlength{\tabcolsep}{1mm}
\begin{tabular}{ccccccc}
\hline
\textbf{Method}  & \textbf{Acc}   & \textbf{R.Acc} & \textbf{F.Acc} & \textbf{Macro-F1} & \textbf{AUC}   & \textbf{AP(fake)} \\ \hline
\textbf{IID}     & 89.64          & 88.44          & 90.85          & 89.64             & 95.52          & 96.68             \\
\textbf{NPR}     & 80.88          & 87.29          & 74.46          & 80.80             & 89.24          & 90.53             \\
\textbf{FreqNet} & 80.34          & 76.81          & 83.86          & 80.31             & 89.31          & 91.58             \\
\textbf{ForAda}  & 84.34          & 76.37          & 92.31          & 84.24             & 94.50          & 95.70             \\
\textbf{AIDE}    & 86.63          & 78.14          & 95.11          & 86.53             & 95.07          & 94.51             \\
\textbf{Effort}  & 91.87          & 87.99          & 95.74          & 91.86             & 98.29          & 98.53             \\
\textbf{PGC}     & 90.37          & 90.22          & 90.53          & 90.37             & 96.41          & 97.06             \\ \hline
\textbf{Ours}    & \textbf{96.54} & \textbf{95.30} & \textbf{97.78} & \textbf{96.54}    & \textbf{99.50} & \textbf{99.58}    \\ \hline
\end{tabular}
\caption{\textbf{OOD evaluation on the GPT-Image-2 subset.}}
\label{tab:ood-gpt-image-2}
\end{table}

\noindent\textbf{Multi-Level Feature Fusion and Classification:}
Different network depths produce visual representations at different levels of abstraction. Low-, middle-, and high-level representations may each contain information useful for real/fake classification, and such information may not be fully retained in the final-layer CLS token. We therefore fuse CLS tokens from multiple depths, allowing the classifier to exploit complementary representations across the network. In Stage 2, we freeze the self-supervised visual backbone $f_{\theta}$ and train only an MLP classifier $m_{\psi}$. Given a full, unmasked image $\mathbf{x}_i$, let the backbone contain $L$ ViT blocks and let $\mathbf{c}_i^{(\ell)}$ denote the CLS token output by block $\ell$:
\begin{equation}
\mathbf{c}_i^{(\ell)}=\left[f_{\theta}^{(\ell)}(\mathbf{x}_i)\right]_{\mathrm{CLS}},\quad
\ell\in\mathcal{S}=\{\ell_{\mathrm{low}},\ell_{\mathrm{mid}},\ell_{\mathrm{high}}\}.
\label{eq:multilevel-cls}
\end{equation}
These layers provide low-, middle-, and high-level representations, respectively. We first $\ell_2$-normalize each CLS token:
\begin{equation}
\bar{\mathbf{c}}_i^{(\ell)}=\frac{\mathbf{c}_i^{(\ell)}}{\|\mathbf{c}_i^{(\ell)}\|_2},\quad \ell\in\mathcal{S}.
\label{eq:cls-normalization}
\end{equation}
The normalized tokens are then concatenated into a fused feature $\mathbf{h}_i$:
\begin{equation}
\mathbf{h}_i=[\bar{\mathbf{c}}_i^{(\ell_{\mathrm{low}})};\bar{\mathbf{c}}_i^{(\ell_{\mathrm{mid}})};\bar{\mathbf{c}}_i^{(\ell_{\mathrm{high}})}].
\label{eq:cls-fusion}
\end{equation}
The MLP produces the real/fake prediction
\begin{equation}
\hat{\mathbf{y}}_i=\mathrm{softmax}\left(m_{\psi}(\mathbf{h}_i)\right),
\label{eq:classification-prediction}
\end{equation}
and is optimized using the cross-entropy loss
\begin{equation}
\mathcal{L}_{\mathrm{cls}}=-\frac{1}{B}\sum_{i=1}^{B}\sum_{c=1}^{2}y_{i,c}\log\hat{y}_{i,c}.
\label{eq:classification-loss}
\end{equation}

\section{Experiment}
\begin{table}
\centering

\small
\setlength{\tabcolsep}{1mm}
\begin{tabular}{ccccccc}
\hline
\textbf{Method}  & \textbf{Acc}   & \textbf{R.Acc} & \textbf{F.Acc} & \textbf{Macro-F1} & \textbf{AUC}   & \textbf{AP(fake)} \\ \hline
\textbf{IID}     & 78.60          & 81.15          & 76.05          & 78.59             & 86.65          & 88.53             \\
\textbf{NPR}     & 81.68          & 84.84          & 78.52          & 81.66             & 89.54          & 91.06             \\
\textbf{FreqNet} & 70.65          & 79.04          & 62.25          & 70.44             & 77.61          & 79.13             \\
\textbf{ForAda}  & 82.49          & 82.44          & 82.55          & 82.49             & 90.50          & 91.87             \\
\textbf{AIDE}    & 73.01          & 72.56          & 73.47          & 73.01             & 81.21          & 83.23             \\
\textbf{Effort}  & 85.17          & 87.32          & 83.01          & 85.16             & 92.98          & 94.16             \\
\textbf{PGC}     & 82.17          & 83.05          & 81.29          & 82.17             & 89.98          & 91.57             \\ \hline
\textbf{Ours}    & \textbf{89.51} & \textbf{90.84} & \textbf{88.19} & \textbf{89.51}    & \textbf{95.36} & \textbf{96.03}    \\ \hline
\end{tabular}
\caption{\textbf{Comparison with state-of-the-art detection baselines.}}
\label{tab:baseline-comparison}
\end{table}

\subsection{Experiment Settings}
\noindent\textbf{Training Setup: }
All experiments are conducted on 8 NVIDIA RTX A6000 GPUs. The MAE decoder comprises 8 ViT blocks. In Stage I, the model is trained for 70 epochs. Specifically, we freeze the visual backbone and optimize only the MAE decoder during the first 30 epochs. For the remaining 40 epochs, we unfreeze the visual backbone and jointly optimize it with the MAE decoder. In this stage, the learning rates are set to $1\times10^{-4}$ for the MAE decoder and $1\times10^{-5}$ for the visual backbone, with a batch size of 4 per GPU. For contrastive learning, we set $G=2$, $K=2$, and the temperature to $\tau=0.1$. During joint optimization, we optimize $\mathcal{L}_{\mathrm{MAE}}+0.1\mathcal{L}_{\mathrm{con}}$. Following the original MAE configuration, we use a masking ratio of $\rho=0.75$, while the number of image patches $P$ and the hidden feature dimension follow the default configuration of each visual backbone. In Stage II, we freeze the visual backbone and train only the MLP classifier for 10 epochs, using a learning rate of $1\times10^{-4}$ and a batch size of 64. Unless otherwise specified, all input images are resized to $224\times224$ during training. For reproducibility, we set the random seed to 42 for all experiments.

\begin{table}
\centering

\small
\setlength{\tabcolsep}{1mm}
\begin{tabular}{ccccccc}
\hline
\textbf{Method}  & \textbf{Acc}   & \textbf{R.Acc} & \textbf{F.Acc} & \textbf{Macro-F1} & \textbf{AUC}   & \textbf{AP(fake)} \\ \hline
\textbf{IID}     & 67.64          & 81.85          & 53.42          & 66.97             & 70.89          & 75.47             \\
\textbf{NPR}     & 73.37          & 83.00          & 63.74          & 73.12             & 79.69          & 83.06             \\
\textbf{FreqNet} & 69.91          & 71.50          & 68.33          & 69.91             & 75.52          & 78.18             \\
\textbf{ForAda}  & 76.59          & 72.47          & 80.71          & 76.55             & 85.19          & 86.61             \\
\textbf{AIDE}    & 59.76          & 56.77          & 62.76          & 59.73             & 62.22          & 60.06             \\
\textbf{Effort}  & 79.43          & 84.04          & 74.83          & 79.39             & 87.62          & 89.52             \\
\textbf{PGC}     & 83.46          & 92.09          & 74.82          & 83.33             & 93.35          & 93.86             \\ \hline
\textbf{Ours}    & \textbf{86.49} & \textbf{95.88} & \textbf{77.09} & \textbf{86.37}    & \textbf{95.61} & \textbf{95.94}    \\ \hline
\end{tabular}
\caption{\textbf{OOD evaluation on the HyperSwapper subset.}}
\label{tab:ood-hyperswapper}
\end{table}
\begin{table}
\centering

\small
\setlength{\tabcolsep}{1mm}
\begin{tabular}{ccccccc}
\hline
\textbf{Method}  & \textbf{Acc}   & \textbf{R.Acc} & \textbf{F.Acc} & \textbf{Macro-F1} & \textbf{AUC}   & \textbf{AP(fake)} \\ \hline
\textbf{IID}     & 59.42          & 98.15          & 20.69          & 52.26             & 60.93          & 68.73             \\
\textbf{NPR}     & 64.65          & 85.43          & 43.88          & 63.06             & 69.54          & 72.63             \\
\textbf{FreqNet} & 60.42          & 60.16          & 60.68          & 60.42             & 63.73          & 61.34             \\
\textbf{ForAda}  & 55.46          & 81.47          & 29.45          & 52.22             & 57.15          & 58.63             \\
\textbf{AIDE}    & 55.85          & 51.90          & 59.79          & 55.78             & 56.17          & 52.76             \\
\textbf{Effort}  & 60.65          & 88.98          & 32.31          & 57.21             & 63.83          & 67.90             \\
\textbf{PGC}     & 64.67          & 94.21          & 35.13          & 61.29             & 76.69          & 78.22             \\ \hline
\textbf{Ours}    & \textbf{80.93} & \textbf{95.44} & \textbf{66.42} & \textbf{80.52}    & \textbf{91.94} & \textbf{92.69}    \\ \hline
\end{tabular}
\caption{\textbf{OOD evaluation on the PersonaLive subset.}}
\label{tab:ood-personalive}
\end{table}

\noindent\textbf{Dataset: }
We construct an identity-disjoint benchmark of 147,900 images from 361,469 successfully rendered and decodable candidates, including 117,900 training, 15,000 validation, and 15,000 test images. To prevent identity leakage, samples linked by any shared person ID are grouped into a connected component and assigned entirely to one split. Within each split, real and fake samples follow an exact $1{:}1$ ratio; fake samples are equally divided among text-to-image generation, expression-driven manipulation, and face swapping; and the five 3D reconstruction methods each account for 20\% of the samples. The same class and forgery-category proportions are maintained within each reconstruction method. Because only complete identity components satisfying these constraints are retained, the final benchmark is smaller than the candidate pool. It contains 13,981 unique identities, with no person IDs or identical rendered images shared across splits.

\noindent\textbf{Baseline: }
We select recent detection models published at top-tier conferences that introduce substantive methodological innovations and can be retrained on our dataset. This selection enables all baselines to be evaluated under a unified training protocol, ensuring a fair comparison. Specifically, the selected baselines include four general-purpose real-versus-fake image detectors: AIDE~\cite{adie}~(ICLR 2025), Effort~\cite{effort}~(ICML 2025), PGC~\cite{pgc}~(ICML 2026), and FreqNet~\cite{freqnet}~(AAAI 2024). We also evaluate three detectors developed for face forgery and deepfake detection: IID~\cite{iid}~(CVPR 2023), ForAda~\cite{ForensicsAdapter}~(CVPR 2025), and NPR~\cite{npr}~(CVPR 2024). For all baselines, we standardize the input resolution to $224\times224$ and follow the best-performing training configuration provided by each method's official implementation for all other hyperparameters. For the main comparison, our model uses DINOv3 ViT-H+/16~\cite{dinov3} and concatenates the CLS tokens from blocks 10, 18, and 23 (zero-based) for Stage II classification.

\noindent\textbf{Evaluation Metrics:}
We report overall accuracy (ACC), class-specific accuracies (Real-ACC and Fake-ACC), the macro-averaged F1 score (Macro-F1), the area under the receiver operating characteristic curve (AUC), and average precision with fake as the positive class (AP$_{\mathrm{fake}}$). All results are reported as percentages(\%).

\begin{figure}
    \centering
    \includegraphics[width=1.0\linewidth]{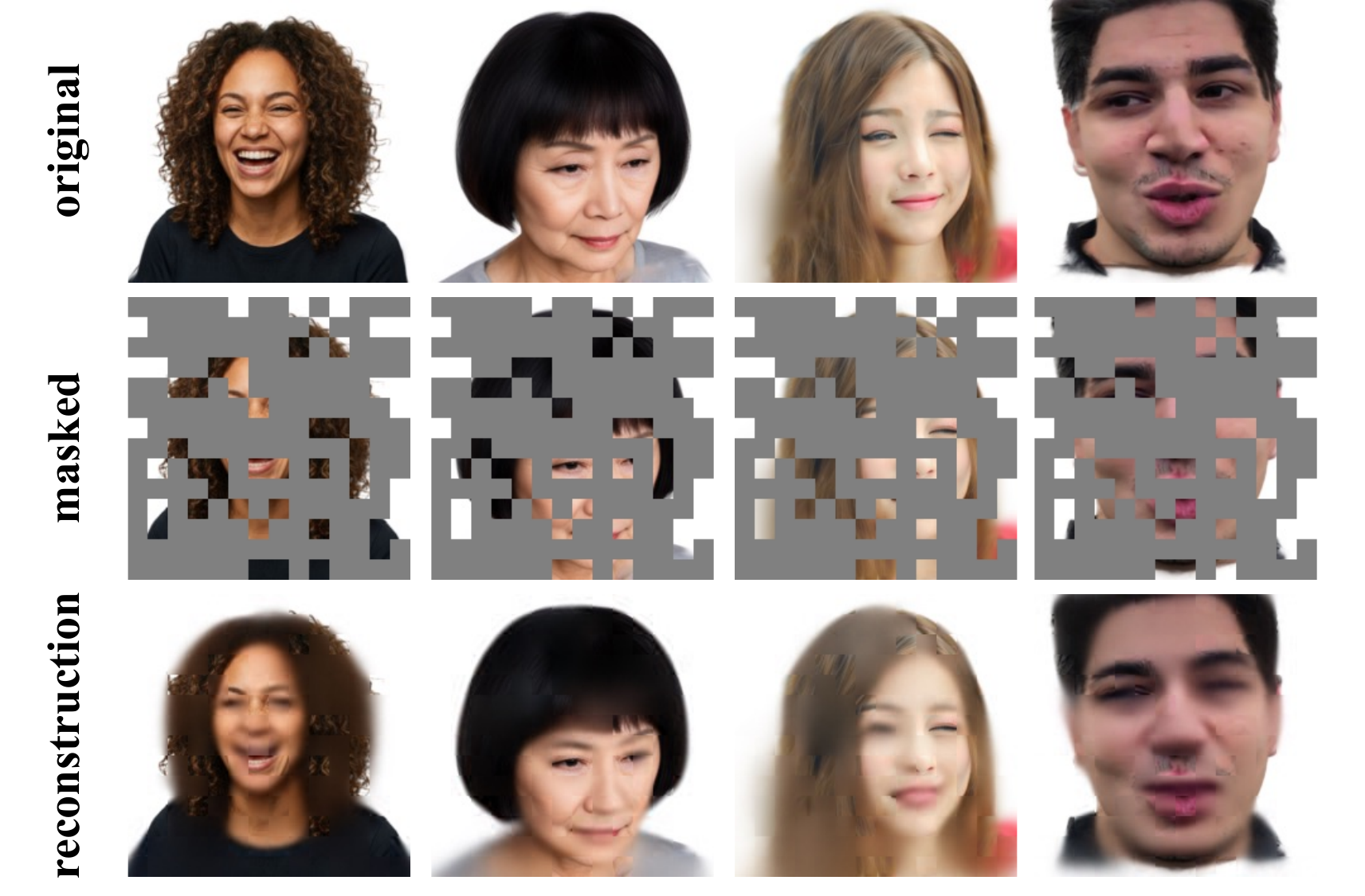}
    \caption{\textbf{Qualitative masked-reconstruction results.} With 75\% of the input patches masked, the MAE recovers coherent facial structure and fine-grained local details in both real and fake 3D Gaussian-rendered heads.}
    \label{fig:mae}
\end{figure}
\begin{table}[]
\centering
\small
\setlength{\tabcolsep}{1mm}
\begin{tabular}{ccccccc}
\hline
\textbf{Method}  & \textbf{Acc}                  & \textbf{R.Acc} & \textbf{F.Acc} & \textbf{Macro-F1} & \textbf{AUC}   & \textbf{AP(fake)} \\ \hline
\textbf{IID}     & 82.55                         & 81.91          & 83.19          & 82.55             & 90.16          & 92.06             \\
\textbf{NPR}     & 81.83 & 88.14          & 75.53          & 81.76             & 87.66          & 90.70             \\
\textbf{FreqNet} & 68.87                         & 72.92          & 64.82          & 68.82             & 76.08          & 77.85             \\
\textbf{ForAda}  & 83.60                         & 89.85          & 77.34          & 83.53             & 90.78          & 92.84             \\
\textbf{AIDE}    & 81.40                         & 89.11          & 73.69          & 81.29             & 87.20          & 90.19             \\
\textbf{Effort}  & 85.46                         & 89.57          & 81.37          & 85.44             & 92.16          & 93.95             \\
\textbf{PGC}     & 83.33                         & 87.64          & 79.03          & 83.30             & 90.26          & 92.59             \\ \hline
\textbf{Ours}    & \textbf{89.24}                & \textbf{91.61} & \textbf{86.87} & \textbf{89.24}    & \textbf{95.62} & \textbf{96.53}    \\ \hline
\end{tabular}
\caption{\textbf{OOD evaluation on the Any3DAvatar subset.}}
\label{tab:ood-any3davatar}
\end{table}

\subsection{Comparison with Baselines}
Tab.~\ref{tab:baseline-comparison} compares our method with recent general-purpose and face-oriented forgery detectors under the unified benchmark protocol. Our method achieves the best result across all evaluation metrics and consistently outperforms the strongest baseline. The gains in both Real-ACC and Fake-ACC further show that the improvement is balanced across the two classes rather than driven by a bias toward either real or fake samples.
Tab.~\ref{tab:split-results} provides a finer-grained comparison across the 2D source categories and 3D reconstruction methods. Our method ranks first in every reported category and reconstruction pipeline, demonstrating consistent performance across the balanced IID benchmark.
For OOD evaluation, we exclude the newest method in each forgery category from training and reserve it for testing: GPT-Image-2 for text-to-image generation, PersonaLive for expression-driven manipulation, , HyperSwapper for face swapping and Any3DAvatar for 3D methods. As shown in Tab.~\ref{tab:ood-gpt-image-2}, Tab.~\ref{tab:ood-hyperswapper}, Tab.~\ref{tab:ood-personalive}, and Tab.~\ref{tab:ood-any3davatar}, our method achieves the best results across all reported metrics on all four held-out subsets, demonstrating stronger generalization to unseen forgery methods and an unseen 3D reconstruction method.

\subsection{Ablative Analyses}\label{sec:ablation}

\noindent\textbf{Masked Autoencoding:}
To assess the effectiveness of masked autoencoding, we remove the MAE decoder and $\mathcal{L}_{\mathrm{MAE}}$ from Stage I while keeping all other components and training settings unchanged. As shown in Tab.~\ref{tab:stage1-ablation}, this removal consistently degrades all evaluation metrics, demonstrating that masked reconstruction provides complementary supervision and contributes to downstream authenticity classification.
Fig.~\ref{fig:mae} shows that, even with 75\% masking, MAE produces visually plausible reconstructions and recovers some personalized facial details in both real and fake samples. This observation suggests that the encoder attends to fine-grained local appearance information retained after 3D Gaussian reconstruction, supporting its usefulness for downstream authenticity classification together with the quantitative ablation.

\begin{table}[]
\centering

\small
\setlength{\tabcolsep}{1mm}
\begin{tabular}{ccccccc}
\hline
\textbf{Strategy}              & \textbf{Acc}   & \textbf{R.Acc} & \textbf{F.Acc} & \textbf{Macro-F1} & \textbf{AUC}   & \textbf{AP(fake)} \\ \hline
\textbf{w/o Stage I}   & 85.79          & 86.39          & 85.20          & 85.79             & 93.18          & 94.14             \\
\textbf{w/o MAE}               & 86.96          & 88.40          & 85.52          & 86.96             & 93.81          & 94.69             \\
\textbf{w/o CL} & 87.18          & 88.56          & 85.80          & 87.18             & 94.16          & 94.85             \\
\textbf{Ours}                  & \textbf{89.51} & \textbf{90.84} & \textbf{88.19} & \textbf{89.51}    & \textbf{95.36} & \textbf{96.03}    \\ \hline
\end{tabular}
\caption{\textbf{Ablation study of Stage I training.} We report results without masked autoencoding (w/o MAE), without multi-view contrastive learning (w/o CL), and without Stage I training (w/o Stage I). All results are reported as percentages.}
\label{tab:stage1-ablation}
\end{table}

\begin{table}[t]
\centering
\small
\setlength{\tabcolsep}{3.5pt}
\begin{tabular}{lcccccc}
\toprule
\multirow{2}{*}{\textbf{Strategy}}
& \multicolumn{2}{c}{\textbf{Intra-Sim} $\uparrow$}
& \multicolumn{2}{c}{\textbf{Inter-Sim} $\downarrow$}
& \multicolumn{2}{c}{\textbf{Sep-Gap} $\uparrow$} \\
\cmidrule(lr){2-3}
\cmidrule(lr){4-5}
\cmidrule(lr){6-7}
& Real & Fake & Real & Fake & Real & Fake \\
\midrule
w/o CL
& 0.7427 & 0.7334
& 0.3499 & 0.3678
& 0.3928 & 0.3656 \\
w/ CL
& \textbf{0.8965} & \textbf{0.8926}
& \textbf{0.3293} & \textbf{0.3470}
& \textbf{0.5672} & \textbf{0.5456} \\
\bottomrule
\end{tabular}
\caption{\textbf{Cross-view representation analysis.} Intra-Sim measures similarity across views of the same 3D head, Inter-Sim across different heads of the same authenticity class, and Sep-Gap is their difference.}
\label{tab:contrastive-similarity}
\end{table}

\begin{figure}[t]
    \centering
    \includegraphics[width=1.0\linewidth]{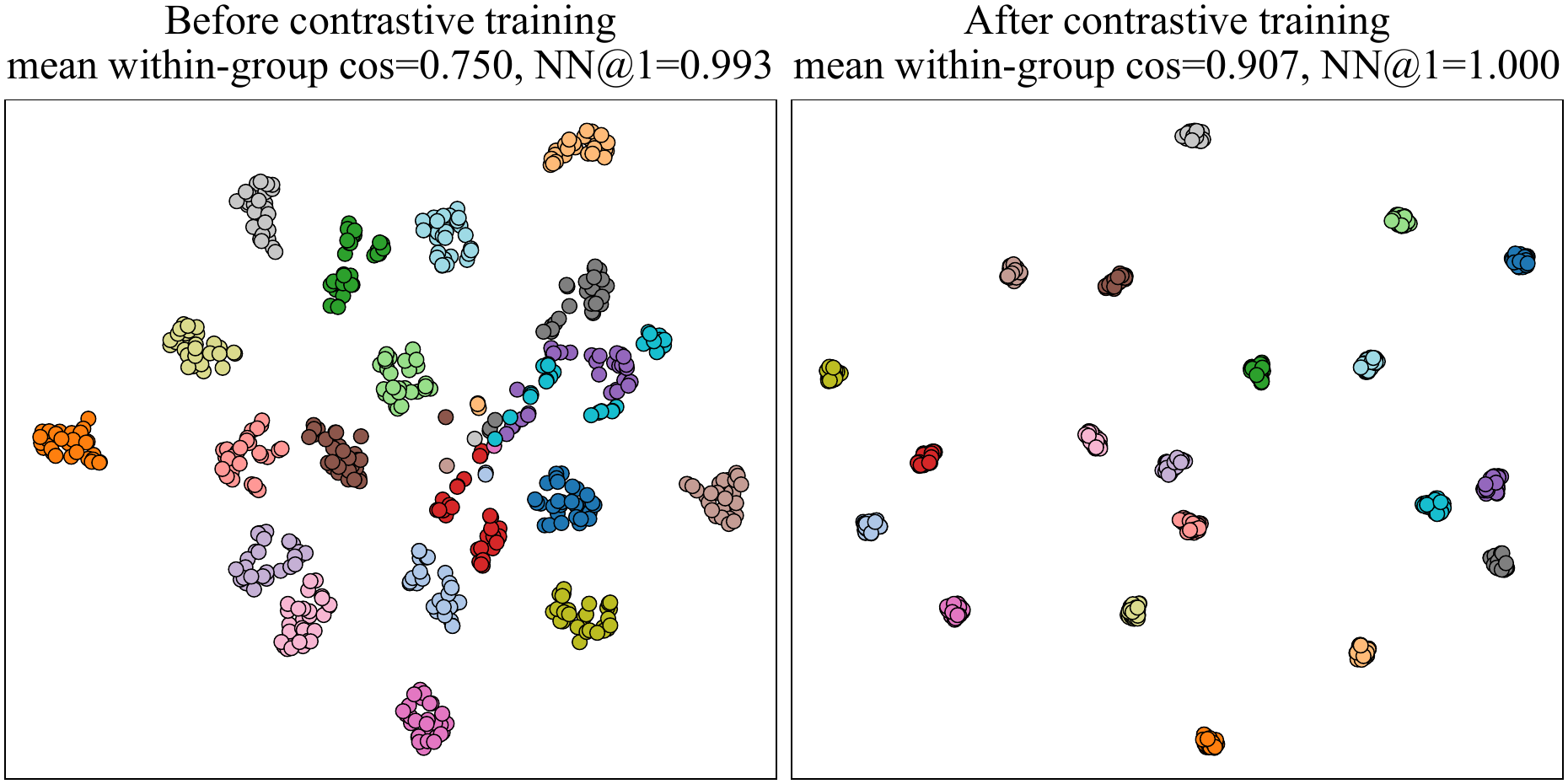}
    \caption{\textbf{t-SNE of multi-view representations} before (left) and after (right) contrastive learning, showing tighter within-head clusters and better separation across identities.}
    \label{fig:contrast}
\end{figure}

\begin{figure}
    \centering
    \includegraphics[width=1.0\linewidth]{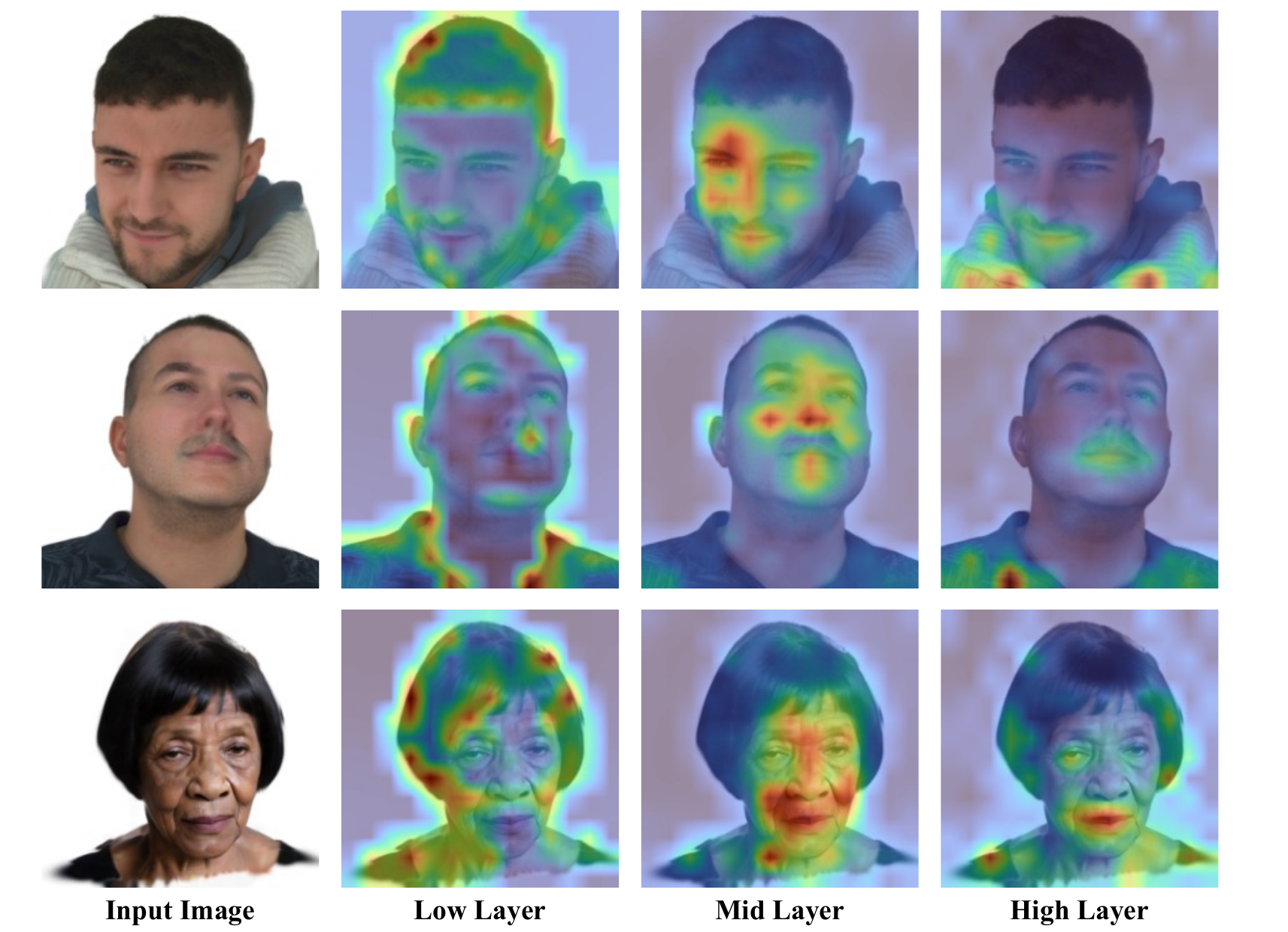}
    \caption{\textbf{CLS-token attention across network depths.} CLS-token attention maps, averaged over all attention heads, for the selected low-, middle-, and high-level blocks. Different depths exhibit distinct spatial response patterns.}
    \label{fig:multilayer}
\end{figure}

\begin{table}[]
\centering
\small
\setlength{\tabcolsep}{0.5mm}
\begin{tabular}{ccccccc}
\hline
\textbf{Strategy}       & \textbf{Acc}   & \textbf{R.Acc} & \textbf{F.Acc} & \textbf{Macro-F1} & \textbf{AUC}   & \textbf{AP(fake)} \\ \hline
\textbf{w/ Low-Layer}   & 68.57          & 69.26          & 67.87          & 68.53             & 75.04          & 75.72             \\
\textbf{w/ Mid-Layer}   & 83.16          & 84.8           & 81.50          & 83.15             & 90.76          & 92.09             \\
\textbf{w/ High-Layer}  & 85.64          & 87.46          & 83.82          & 85.64             & 92.68          & 93.80             \\
\textbf{w/ Last-Layer}  & 84.31          & 85.93          & 82.69          & 84.31             & 91.62          & 92.95             \\
\textbf{w/ Multi-Layer} & \textbf{89.51} & \textbf{90.84} & \textbf{88.19} & \textbf{89.51}    & \textbf{95.36} & \textbf{96.03}    \\ \hline
\end{tabular}
\caption{\textbf{Stage II ablation of CLS token selection.} We compare the average results of single-layer classifiers in the low-, middle-, and high-depth ranges, the last-layer classifier, and low--middle--high CLS token fusion.}
\label{tab:stage2-multilevel-ablation}
\end{table}

\noindent\textbf{Multi-View Contrastive Learning:}
To analyze the learned representation, we compute cosine similarity between final-layer CLS tokens from approximately 13,000 images in 3,800 balanced multi-view groups, forming about 52,000 positive and negative pairs. Tab.~\ref{tab:contrastive-similarity} shows that contrastive learning increases Sep-Gap mainly by improving intra-head similarity, confirming stronger cross-view consistency. Removing the contrastive branch and objective consistently degrades performance (Tab.~\ref{tab:stage1-ablation}). Consistently, the t-SNE visualization of 20 balanced identity groups (10 views each) in Fig.~\ref{fig:contrast} shows more compact same-head clusters and clearer inter-group separation.

\noindent\textbf{Multi-Level CLS Token Fusion:}
To examine how the classification utility of CLS tokens varies with network depth, we freeze the Stage I visual backbone and train an identical Stage II MLP classifier using the CLS token from each Transformer block separately. We group the blocks into low-, middle-, and high-depth ranges and report the mean performance of the corresponding single-layer probes, while reporting the final block separately. As shown in Tab.~\ref{tab:stage2-multilevel-ablation}, CLS tokens across different depths all contain information useful for authenticity classification, while multi-level fusion consistently outperforms the single-layer averages and the last-layer classifier.
To inspect how the spatial responses of CLS tokens vary across depth, we average the CLS-token attention weights over all attention heads for each visualized block. As shown in Fig.~\ref{fig:multilayer}, different layers exhibit distinct attention regions, providing qualitative evidence that their spatial responses are not identical.
Based on layer-wise probing and attention diversity, we select blocks 10, 18, and 23 (zero-based) as representative low-, middle-, and high-level layers and concatenate their normalized CLS tokens for Stage II classification.

\noindent\textbf{Visual Backbone:}
Tab.~\ref{tab:backbone-ablation} examines the effects of backbone generation and model capacity. At a comparable scale, DINOv3 outperforms DINOv2, indicating the benefit of newer pretrained representations. Within DINOv3, performance consistently improves as model capacity increases, with ViT-H+/16 achieving the best overall results and therefore serving as our default backbone. Importantly, our framework still outperforms the strongest baseline when using either the older DINOv2 backbone or the smallest DINOv3 variant. This shows that stronger backbones further improve performance, while the gains of our method cannot be attributed solely to backbone scale or recency.
\begin{table}[]
\centering

\small
\setlength{\tabcolsep}{0.3mm}
\begin{tabular}{ccccccc}
\hline
\textbf{Visual Backbone}  & \textbf{Acc}   & \textbf{R.Acc} & \textbf{F.Acc} & \textbf{Macro-F1} & \textbf{AUC}   & \textbf{AP(fake)} \\ \hline
\textbf{DINOv2 ViT-L/14}  & 86.92          & 88.20          & 85.64          & 86.92             & 93.35          & 94.40             \\
\textbf{DINOv3 ViT-B/16}  & 87.19          & 90.07          & 84.31          & 87.18             & 94.38          & 95.14             \\
\textbf{DINOv3 ViT-L/16}  & 88.49          & 90.13          & 86.85          & 88.49             & 95.05          & 95.80             \\
\textbf{DINOv3 ViT-H+/16} & \textbf{89.51} & \textbf{90.84} & \textbf{88.19} & \textbf{89.51}    & \textbf{95.36} & \textbf{96.03}    \\ \hline
\end{tabular}
\caption{\textbf{Comparison of visual backbone versions and model scales for source-face authenticity detection.} DINOv3 ViT-H+/16 achieves the best performance across all metrics.}
\label{tab:backbone-ablation}
\end{table}

\section{Conclusion}
\noindent\textbf{Summary: }We introduce the first large-scale dataset and systematic benchmark for real/fake 3D Gaussian head detection, together with a two-stage detector that learns detail-preserving, cross-view-consistent representations and fuses multi-level CLS tokens for classification. Our method consistently outperforms existing detectors.

\noindent\textbf{Limitations: }Expression-driven forgeries remain challenging, with an accuracy of 76.24\%. Future work will expand this category for stronger detection.

\bibliography{aaai2027}


\end{document}